\documentclass[11pt,a4paper]{article}
\usepackage[hyperref]{acl2020}
\usepackage{times}
\usepackage{latexsym}

\usepackage{microtype}

\usepackage{url}

\aclfinalcopy 

\usepackage{amsfonts, amsmath, amssymb, amsthm}
\usepackage{empheq}
\DeclareMathOperator*{\argmax}{arg\,max}
\DeclareMathOperator*{\argmin}{arg\,min}
\usepackage{algorithm}
\usepackage{algcompatible}
\usepackage{multirow}
\usepackage{appendix}

\usepackage{pgfplots}
\pgfplotsset{compat=1.7}
\usepackage{pgfplotstable}
\usepackage{booktabs}
\usepackage{array}
\usepackage{colortbl}
\usepackage{graphicx}
\usepackage{subcaption}
\usepackage{float}
\newcommand{\com}[1]{}

\pgfplotstableset{
  every head row/.style={before row=\toprule,after row=\midrule},
  every last row/.style={after row=\bottomrule},
  fixed,precision=2,
}

\title{The Structured Weighted Violation MIRA}

\author{
  Dor Ringel \quad Rotem Dror \quad Roi Reichart \\
  Faculty of Industrial Engineering and Management, Technion, IIT \\
  {\tt \{dorringel@cs|rtmdrr@campus|roiri@ie\}.technion.ac.il} \\
}

\date{}

\begin{document}
\maketitle
\begin{abstract}
We present the \textit{Structured Weighted Violation MIRA (SWVM)}, a new structured prediction algorithm that is based on an hybridization between MIRA \citep{crammer2003ultraconservative} and the structured weighted violations perceptron (SWVP) \citep{dror2016structured}. We demonstrate that the concepts developed in \citep{dror2016structured} combined with a powerful structured prediction algorithm can improve performance on sequence labeling tasks. In experiments with syntactic chunking and named entity recognition (NER), the new algorithm substantially outperforms the original MIRA as well as the original structured perceptron and SWVP.\footnote{Our code is available at \tt https://github.com/dorringel/SWVM}
\end{abstract}

\section{Introduction}
\label{sec:intro}

Structured prediction (SP) algorithms constitute a cornerstone of problem-solving in natural language processing (NLP). Even in the neural network (NN) era, linear SP algorithms still play a key role. In some works, they were fed with features learned by an NN, and learn a classifier with these features. For example, \citep{weiss2015structured} used the structured perceptron \citep{collins2002discriminative} for training a transition based dependency parser with NN-based features. In other cases, linear SP algorithms do not fall short of NN methods even with manually crafted features (e.g., \citep{goyal2016joint,goyal2016posterior,sharaf2017structured}). Further developing the SP methodology is hence of great importance.

The starting point of this paper is the structured weighted violation perceptron (SWVP) algorithm \citep{dror2016structured}, henceforth DR16, see Algorithm~\ref{alg:SWVP}). SWVP is a generalization of the Collins structured perceptron (CSP), based on the concept of \textit{violations} \citep{huang2012structured}.

In DR16, SWVP substantially outperformed CSP in synthetic data experiments. However, in the real data scenario, when the two algorithms train the TurboParser \citep{martins2013turning} on the data from the CoNLL-2007 shared task on multilingual dependency parsing \citep{nivre2007conll}, they perform similarly.

In this paper, we show that the ideas developed in DR16 can lead to substantial performance gains in real-world NLP tasks when incorporated into the MIRA algorithm (\citep{crammer2003ultraconservative}, see Algorithm~\ref{alg:MIRA}). 
We call our new algorithm SWVM for \textit{Structured Weighted Violations MIRA}. We experiment with three Named Entity Recognition (NER) datasets and one syntactic chunking setup and demonstrate substantial performance gains over the original MIRA as well as the original CSP and SWVP.

\section{From SWVP to SWVM}

SWVP is based on a modification of the parameter update rule of CSP. In CSP, the update rule considers the difference between the gold label, $y$, of a training example, $x$, and the inferred label, $y^*$, of that example. In SWVP, in contrast, the update rule (line 11 of Algorithm~\ref{alg:SWVP}) considers the difference between $y$ and a set of labels derived from the inferred label $y^*$. The SWVP update rule is formed through two decisions that define the derived labels. These decisions are made each time the update rule is employed.

\begin{algorithm}[tb]
    \caption {The SWVP algorithm. $\mathcal{Y}(x^i)$ is the set of candidate labels of the input example $x^i$.}
	\label{alg:SWVP}
	\begin{algorithmic}[1]
    \small
		\STATEx {\bfseries Input:} data $D = \{x^i,y^i\}_{i=1}^{n}$, feature mapping $\Phi$
		\STATEx {\bfseries Output:} parameter vector $\textbf{w}\in \mathbb{R}^d$
		\STATEx {\bfseries Define:} $\Delta\Phi(x,y,z) \triangleq \Phi(x,y) - \Phi(x,z)$
		\STATE Initialize $\textbf{w}=0$.
		\REPEAT
		\FOR{each $(x^i,y^i)\in D$}
		\STATE $y^{*}=\argmax\limits_{y'\in \mathcal{Y}(x^i)}\textbf{w}\cdot\Phi(x^i,y')$
		\IF{$y^{*} \neq y^i$}
        \STATE {\bfseries Define:} $JJ_{x^i} \subseteq 2^{[L_{x^i}]}$
		\FOR{$J \in JJ_{x^i}$}
		\STATE {\bfseries Define:} $m^J$ s.t. $m_k^J=\begin{cases}
		y^{*}_k & k\in J \\
		y^i_k & else
		\end{cases}$
		\ENDFOR
		\STATE 	$\gamma = \textsc{SetGamma}()$
		\STATE $\textbf{w} = \textbf{w}+\sum\limits_{J \in JJ_{x^i}}\gamma(m^J)\Delta\Phi(x^i,y^i,m^J)$
		\ENDIF
		\ENDFOR
		\UNTIL{Convergence}
	\end{algorithmic}
\end{algorithm}

The first decision is on the set of modification templates, denoted with $JJ_{x}$. Each template, $J \in JJ_{x}$, generates a variant of the inferred label $y^{*}$, denoted with $m^{J}$, that is used in the update rule. $m^J$ is identical to $y^{*}$ in all the indexes of $J$, and to $y$, the gold label, otherwise. For example, in the NER task, for the sentence \textit{Moses runs marathons} with the gold label $y=[Person,None,None]$, the erroneous inferred label $y^*=[Person,Person,Person]$ and the template $J=\{3\}$, we get the new derived label $m^J=[Person,None,Person]$.

The second decision is the choice of a weight function, \textsc{SetGamma}, that allocates a weight, $\gamma(m^J)$, for each template, $J\in JJ_x$.
The weight function is designed to balance between violating and non-violating labels. A violating label (or a \textit{violation}, \citep{huang2012structured}) is a label to which the current model assigns higher score than to the gold label. That is, $m^J$ is a violating assignment if $\textbf{w}\cdot[\Phi(x,y)-\Phi(x,m^J)]\le 0$. Following, $m^J$ is a non-violation if $\textbf{w}\cdot[\Phi(x,y)-\Phi(x,m^J)] > 0$.

DR16 proved that for any set of modification templates $JJ_{x}$, if the $\gamma(m^J)$ weights generated by \textsc{SetGamma} for each $J \in JJ_{x}$ respects two conditions, then SWVP converges to a separating hyperplane for any linearly separable training set. The conditions are:

\begin{enumerate}
\small
\item $\sum_{J\in JJ_x} \gamma(m^J) = 1$, $\gamma(m^J) \ge 0, \forall J\in JJ_x.$
\item $\textbf{w}\cdot\sum\limits_{J\in JJ_x}\gamma(m^J)\Delta\Phi(x,y,m^J) \leq 0.$
\end{enumerate}

In addition, they proved that CSP is a special case of SWVP and that the mistake and generalization bounds of SWVP are tighter than those of CSP. Given a feature representation $\Phi(x,y)$, the update rule of SWVP is (line 11 of Algorithm \ref{alg:SWVP}):
\[\textbf{w} = \textbf{w} + \sum_{J\in JJ_x} \gamma(m^J)\cdot [\Phi(x,y)-\Phi(x,m^J)].\]

The idea of using variants of the inferred label or the gold label in the parameter update rule of learning algorithms (CSP and others) has been explored in other works as well  \citep{sontag2010more,huang2012structured}. However, their motivation was mainly speeding up the inference step rather than improving the predictions of the model.

\begin{algorithm}[tb]
	\caption{The Margin-Infused Relaxed Algorithm (MIRA). $\mathcal{L}$ is the hamming loss.}
	\label{alg:MIRA}
	\begin{algorithmic}[1]
    \small
		\STATEx {\bfseries Input:} data $D = \{x^i,y^i\}_{i=1}^{n}$, feature mapping $\Phi$
		\STATEx {\bfseries Output:} parameter vector $\textbf{w}\in \mathbb{R}^d$
		\STATEx {\bfseries Define:} $\Delta\Phi(x,y,z) \triangleq \Phi(x,y) - \Phi(x,z)$
		\STATE Initialize $\textbf{w}^{(0)}=0, j=0$.
        \FOR{$t=1,\ldots,T$}
        \FOR{each $(x^i,y^i)\in D$}
		\STATE $\textbf{w}^{(j+1)} = \argmin_{\textbf{w}} \|\textbf{w}-\textbf{w}^{(j)}\|^2$ s.t. 
        \STATE $\qquad \textbf{w}\cdot\Delta\Phi(x^i,y^i,y) \ge \mathcal{L}(y^i,y) ,\forall y\in \mathcal{Y}(x^i)$
        \ENDFOR
        \ENDFOR
        \STATE $\textbf{w} = \sum_{j=1}^{nT} \frac{\textbf{w}^{(j)}}{nT}$
	\end{algorithmic}
\end{algorithm}

In this paper, we integrate the concepts developed in DR16 to MIRA. MIRA was first proposed for multi-class classification \citep{crammer2003ultraconservative} and then extended to SP \citep{taskar2004max} and has demonstrated strong performance in a variety of structured NLP tasks \citep{mcdonald2005online,watanabe2007online, chiang2008online, bohnet2009efficient,kummerfeld2015empirical}. We integrate the ideas of modification templates and the \textsc{SetGamma} function as proposed in DR16 to the constraints in the optimization problem solved by the MIRA parameter update rule (Figure \ref{fig:SWVM}) and get the Structured Weighted Violation MIRA (SWVM). Just like SWVP, SWVM depends on the definition of $JJ_{x}$  and of \textsc{SetGamma}, which we discuss below.

\begin{figure}
\begin{equation*}
\boxed{
\begin{aligned}
& \underset{\textbf{w}}{\text{minimize}}
\|\textbf{w}-\textbf{w}^{(t)}\|^2 \\
& \text{subject to}\\
& \textbf{w}\cdot\sum_{J\in JJ_{x^i}} \gamma(m^J)\Delta\Phi(x^i,y^i,m^J)\ge \mathcal{L}(y^i,y^*)\\
& \sum_{J\in JJ_{x^i}} \gamma(m^J) = 1\\
& \gamma(m^J) \ge 0, \forall J\in JJ_{x^i}
\end{aligned}}
\end{equation*}
\caption{The SWVM update rule. The input weight vector is $\textbf{w}^{(t)}$. Here we show the constraints only with respect to $y^*$: the top scoring label according to $w^{(t)}$. Like in the original MIRA (Alg. \ref{alg:MIRA}) we can add constraints w.r.t the top K labels according to $w^{(t)}$, where $K \in \{1, \ldots, \mathcal{Y}(x^i) \}$. All other notation is as in Alg. \ref{alg:MIRA}.}
\label{fig:SWVM}
\end{figure}

\section{SWVM Variants}
\label{sec:variations}

In this section, we discuss the two implementation details of SWVM: the definition of the modification template set, $JJ_{x}$, and of the weighting function \textsc{SetGamma}. We note that we cannot show that the theoretical properties of SWVP, as proved by DR16, hold for SWVM. Instead, we explore heuristic ways to integrate their ideas into the MIRA algorithm. In Section~\ref{sec:res}, we show that SWVM outperforms CSP, SWVP, and MIRA in two sequence labeling tasks.

\paragraph{$JJ_{x}$ selection} The set of modification templates for an input example $x$ can be very large. For example, in a tagging task where each word in the input sentence is assigned a single label, there are  $2^{Length(x)}$ possible templates, where $Length(x)$ is the number of words in the input sentence $x$. We consider only the modification templates considered in DR16, i.e. templates of size 1, that indicate a change in a single index ($J=\{i\}, i=1,\ldots,Length(x)$) 
of the gold label $y$. 
%

\paragraph{\textsc{SetGamma}} 
We experiment with several heuristics. Weighted Margin (WM, see below) was proposed in DR16, while the rest are proposed here for the first time:\footnote{We also experimented with the other \textsc{SetGamma} function proposed in DR16: Weighted Margin Rank (WMR), but its results were consistently worse than WM. We hence do not describe it here.} 

\paragraph{\textbf{Uniform.}} Each template gets the same weight.
\[\gamma(m^J) =\frac{1}{|JJ_x|} \]

\paragraph{\textbf{Weighted Margin (WM).}}  Each $m^J$ is given a weight proportional to the violation it causes: the larger is the positive difference between the score of $m^J$ and the score of $y$ according to the current model, the larger is the weight:
\[\gamma(m^J)=\frac{|\min\{\textbf{w}\cdot\Delta\Phi(x,y,m^J),0\}|}{\sum\limits_{{\acute{J}}\in JJ_x}|\min\{\textbf{w}\cdot\Delta\Phi(x,y,m^{\acute{J}}),0\}|}\]

\paragraph{\textbf{Softmin.}} We derive this heuristic from WM. Instead of using the violation value as a weight, we use its exponent. This means that stronger violations (i.e., more negative values of $\textbf{w}\cdot\Delta\Phi(x,y,m^J)$) get exponentially larger weights.
\[\gamma(m^J)=\frac{\exp^{-\textbf{w}\cdot\Delta\Phi(x,y,m^J)}}{\sum\limits_{{\acute{J}}\in JJ_x}\exp^{-\textbf{w}\cdot\Delta\Phi(x,y,m^{\acute{J}})}}\]

\paragraph{\textbf{Optimization}.} We solve an optimization problem for determining the $\gamma$ values. This problem aims to find the $\gamma$ weights that minimize the score of the violations according to the current parameter vector $\textbf{w}$. It is formalized as a maximum function since the violation value is negative.

\begin{align*}
& \underset{\gamma}{\text{maximize}} \quad \textbf{w}\cdot\sum_{J\in JJ_x}\gamma(m^J)\Phi(x,y,m^J)\\
& \text{subject to} &\\
&\quad \sum_{J\in JJ_x}\gamma(m^J)=1,\gamma(m^J)\ge 0,\forall J\in JJ_x\\
& \quad \textbf{w}\cdot\sum_{J\in JJ_x}\gamma(m^J) \Delta \Phi(x,y,m^J)\le 0
\end{align*}

Following DR16, we also consider the \textit{aggressive} approach to the above weighting schemes. In this approach 
modification templates that do not yield violations are excluded from $JJ_x$ before the weights are computed by the weighting schemes.

%
%
%

In the next section we describe our sequence labeling experiments. 


\section{Experiments and Results}
\label{sec:res}


\begin{table*}[ht]
\small
	\centering
	\begin{tabular}{|c|c|c|c|c|c|c|c|c|c|c|c|c|}
		\hline
		\multirow{3}{*}{Alg.} & \multicolumn{9}{c|}{NER}                                                                           & \multicolumn{3}{c|}{Chunking}   \\ \cline{2-13} 
		& \multicolumn{3}{c|}{JNLPBA}     & \multicolumn{3}{c|}{BC2GM}      & \multicolumn{3}{c|}{CoNLL2002} & \multicolumn{3}{c|}{CoNLL2000}  \\ \cline{2-13} 
		& P     & R     & F1              & P     & R     & F1              & P     & R     & F1             & P     & R     & F1              \\ \hline
		SWVM                        & 63.78 & 70.40 & \textbf{66.41*} & 82.71 & 54.68 & \textbf{65.59*} & 82.95 & 73.98 & \textbf{78.11} & 93.37 & 93.05 & \textbf{93.21*} \\ \hline
		MIRA                        & 60.48 & 72.68 & 65.92           & 82.56 & 48.61 & 60.62           & 79.92 & 74.45 & 76.87          & 92.64 & 92.34 & 92.49           \\ \hline
		SWVP                        & 50.20 & 46.64 & 48.19           & 73.00 & 44.60 & 54.6            & 47.53 & 53.59 & 49.58          & 90.65 & 89.64 & 90.15           \\ \hline
		CSP                         & 68.26 & 55.63 & 56.89           & 74.24 & 48.58 & 51.05           & 83.32 & 71.22 & 76.56          & 92.83 & 92.47 & 92.66           \\ \hline
	\end{tabular}
\caption{Results. R stands for recall, P for precision. Statistical significant cases (computed for F1 only) are marked with *. Notice that every measure is averaged over five folds, so (averaged) F1 is not the harmonic mean of the (averaged) R and the (averaged) P.}
\label{tab:results}
\end{table*}

\paragraph{Tasks and Models}

\footnote{Links to the code and data are in the appendix} We consider two sequence labeling tasks: NER and syntactic chunking, as well as four algorithms: SWVM, MIRA, SWVP and CSP.
We implement the SWVM and SWVP algorithms within the Penn StructLearn software package
\citep{mcdonald2006penn} integrated with MALLET \citep{mccallum2002mallet}. 
CSP and MIRA are already implemented in the package. For both tasks our mode is linear chain CRF \citep{lafferty2001conditional} with trinary potentials defined over a standard set of word and tag based features. The full list of features is provided in the appendix.

\paragraph{NER} We experiment with three datasets: (1) the Spanish dataset of the CoNLL2002 shared task on language-independent NER \citep{tjong2002introduction} with 4 NEs: person, location, organization and miscellaneous; (2) the BC2GM corpus  consisting of 20,000 sentences from biomedical publications annotated for mentions of genes \citep{smith2008overview}; and (3) the JNLPBA corpus \citep{kim2004introduction}, based on the GENIA corpus \citep{ohta2002genia}, consisting of 2,404 biomedical abstracts annotated for mentions of 5 NEs: cell line, cell type, DNA, RNA, and protein.

\paragraph{Chunking} We experiment with the dataset of the CoNLL2000 shared task on syntactic chunking \citep{tjong2000introduction}, consisting of the Wall Street Journal Sections 15-18 and 20 of the Penn Treebank \citep{marcus1993building}.

\paragraph{Evaluation}
 
We compute micro-averaged Recall, Precision and F1 scores: where true-positive, false-positive and false-negative values are computed for each entity mention in the NER datasets and each chunk in the chunking dataset. 
We employ a 5-fold cross-validation protocol for each task and dataset. In each setup the hyper-parameters are tuned on development data and the best configuration is employed to the test data. We report the average evaluation measure across the five folds. For statistical significance we employ the T-test with replicability analysis \citep{dror2017replicability} to check whether the F1 differences between the best and the second-best models for each dataset are significant. More details about cross-validation, hyper-parameter tuning and statistical significance are in the appendix.





\paragraph{Results}

Table~\ref{tab:results} presents our results. In all four setups, SWVM is the best performing algorithm. Considering the F1 gaps from the second best algorithm (MIRA or CSP), we get a maximum gap of 4.97 and an averaged gap of 1.81.

The \textsc{SetGamma} functions of SWVM, as tuned on development data, are: JNLPBA: \textit{Uniform}, BC2GM: \textit{Softmin}; CoNLL2002: \textit{Otimization}; and CoNLL2000: \text{Optimization}.\footnote{Complete hyper-parameter configurations for all four algorithms are provided in the appendix.} 
That is, in all four cases it is one of our novel \textsc{SetGamma} functions that provides the best result. The F1 gap between the best SWVM configuration and SWVM with \textsc{SetGamma = WM} is up to 0.39 with an average of 0.19 (WM is the \textsc{SetGamma} function proposed by DR16, not shown in the table).


Interestingly, in none of the 8 cases it was SWVP - the algorithm from which we borrow the ideas that yield the SWVM algorithm from MIRA - that is second best. In fact, in 7 out of 8 cases SWVP was outperformed by all other algorithms, often by large gaps. This further emphasizes the contribution of our paper. While the ideas of DR16 are theoretically sound, their practical value is limited, at least with the \textsc{SetGamma} functions and the modification templates proposed in DR16 and here. Here we show that the ideas of DR16 do have practical value, when integrated into MIRA.
 
\section{Conclusions}
\label{sec:conclusions}
We presented the SWVM algorithm, a new structured prediction algorithm derived from MIRA using the ideas presented in DR16 for the CSP algorithm. We further proposed three new \textsc{SetGamma} functions and experimentally demonstrated their value.
While we do not provide theoretical guarantees for SWVM, its experimental results on two sequence labeling tasks, NER and syntactic chunking, are promising. 

Future work includes theoretical analysis of SWVM. On the practical side, we hope to find improved \textsc{SetGamma} functions and modification templates, ideally automating this process. Finally, we hope to be able to integrate SWVM with non-linear deep neural networks, to get the best of both worlds.



\section*{Acknowledgments}
We would like to thank Raz Fakterman and Elad Kravi for their contribution to the programmatic endeavors of this work.

\bibliographystyle{acl_natbib}
\bibliography{acl2020}

\begin{table*}[t]
\centering
\begin{tabular}{|c|c|c|c|c|}
\hline
\multirow{3}{*}{Alg.} & \multicolumn{3}{c|}{NER} & \multicolumn{1}{c|}{Chunking} \\ \cline{2-5} 
 & JNLPBA & BC2GM & CoNLL2002 & CoNLL2000 \\ \cline{1-5} 
SWVM & agg., uniform, 1 &  agg., softmin, 1  & agg., opt., 3 &  agg., opt., 5 \\ \hline
MIRA & k=1  & k=5 & k=3& k=3 \\ \hline
SWVP & agg., opt., 1 &  agg., wm, 1 & agg., softmin, 3 & agg., uniform, 1    \\ \hline
\end{tabular}%
\caption{Best hyper-parameter configuration (agrressive/balanced, setGamma, K). agg. stands for \textit{aggressive}; opt. stands for \textit{optimization}. The CSP algorithm does not have hyper-parameters.}
\label{tab:best_config}
\end{table*}

\appendix

\section*{Appendix}
\section{Code and Data}

We implement the SWVM and SWVP algorithms within the Penn StructLearn software package \citep{mcdonald2006penn}\footnote{\tt http://webee.technion.ac.il/people/
koby/code-index.html}, and integrated with MALLET \citep{mccallum2002mallet}.\footnote{\tt http://mallet.cs.umass.edu} CSP and MIRA are already implemented in the package.



Table \ref{tab:dataset_urls} shows the URL from which each of the datasets are retrieved.







\begin{table*}[b]
\centering
\begin{tabular}{|c|c|c|}
\hline
Name & Task & URL \\
\hline
BC2GM & NER & \tt https://github.com/spyysalo/bc2gm-corpus \\
\hline
CoNLL2002 & NER & \tt http://lcg-www.uia.ac.be/conll2002/ner \\
\hline
JNLPBA & NER & \tt https://github.com/spyysalo/jnlpba \\
\hline
CoNLL2000 & Chunking & \tt https://www.clips.uantwerpen.be/conll2000/chunking \\
\hline
\end{tabular}%
\caption{URLs from which each dataset have been retrieved.}
\label{tab:dataset_urls}
\end{table*}

\section{Features}

For both the NER and the chunking tasks, our model is linear chain CRF \citep{lafferty2001conditional} with trinary potentials defined over a standard set of word and tag based features. The full list of features, when considering a word $w$ at position $i$, is as follows:\\

Unigrams (6 feature templates):\\ \\
$w[i],w[i-1],w[i+1],t[i],t[i-1],t[i+1]$. \\

Bigrams (9 feature templates):\\ \\
$(w[i],w[i-1]),(w[i],w[i+1]),(w[i-1],w[i+1]),\\ \\
(t[i],t[i-1]),(t[i],t[i+1]),(t[i-1],t[i+1]),\\ \\
(w[i],t[i-1]),(w[i],t[i]),(w[i],t[i+1])$. \\

Trigrams (4 feature templates):\\ \\
$(t[i-1],t[i], t[i+1]), (w[i],t[i], t[i+1]), \\ \\
(w[i],t[i-1], t[i+1]),  (w[i],t[i], t[i-1])$.\\

\section {Cross-validation and Hyper-parameter Tuning}

In all experiments, we first unify the original train/dev/test split if exists and run a 5-fold cross-validation protocol on the unified set (80\% is randomly sampled for training, 10\% for development and 10\% for test). We tune the hyper-parameters on the development data of each fold, according to micro-averaged F1, selecting the configuration that led to the best average F1 score across the development data sets of the five folds.
 
All the algorithms converged by up to 15 iterations. The aggressive approach was dominant in all development data experiments. For MIRA and SWVM, instead of going over all possible labels, $y\in \mathcal{Y}(x)$, when solving the optimization problem of the parameter update rule, we only consider the $K$-best labels for each example (for $K=1,3,5$). 

The best hyper-parameter configurations for each of the setups are provided in Table \ref{tab:best_config}.

\section {Statistical Significance}

We compute the statistical significance in the following manner. We treat the different folds as dependent datasets, and calculate the p-value for each fold separately. Then, we follow the guidelines from \citep{dror2017replicability}\footnote{\tt https://github.com/rtmdrr/
replicability-analysis-NLP} to perform replicability analysis for dependent datasets with K-Bonferroni. Only if this analysis considers one algorithm to be better than the other for all five folds, we consider the difference between the algorithms to be significant.

\end{document}